\documentclass[letterpaper]{article} 
\usepackage{aaai2026}  
\usepackage{times}  
\usepackage{helvet}  
\usepackage{courier}  
\usepackage[hyphens]{url}  
\usepackage{graphicx} 
\usepackage{natbib}  
\usepackage{caption} 
\usepackage{algorithm}
\usepackage{algorithmic}
\usepackage{amsmath} 
\usepackage{amssymb} 
\usepackage{bm}      
\usepackage{newfloat}
\usepackage{listings}
\usepackage{xcolor}
\DeclareMathOperator*{\argmax}{arg\,max}

\DeclareCaptionStyle{ruled}{labelfont=normalfont,labelsep=colon,strut=off} 
\floatstyle{ruled}
\newfloat{listing}{tb}{lst}{}
\floatname{listing}{Listing}
\usepackage{multirow}

\usepackage{booktabs} 
\usepackage{multirow} 

\title{Dynamic Gaussian Scene Reconstruction from Unsynchronized Videos}
\author{
    Zhixin Xu\textsuperscript{\rm 1}\equalcontrib, 
    Hengyu Zhou\textsuperscript{\rm 1}\equalcontrib, 
    Yuan Liu\textsuperscript{\rm 2}, 
    Wenhan Xue\textsuperscript{\rm 1}, 
    Hao Pan\textsuperscript{\rm 1}, 
    Wenping Wang\textsuperscript{\rm 3}, 
    Bin Wang\textsuperscript{\rm 1,\,4}\thanks{Corresponding Author}
}
\affiliations{

    \textsuperscript{\rm 1}School of Software, Tsinghua University
    
    \textsuperscript{\rm 2}Hong Kong University of Science and 
    Technology

    \textsuperscript{\rm 3}Texas A\&M University

    \textsuperscript{\rm 4}Beijing National Research Center for Information Science and Technology
    
    \{xuzx23, zhouhy22, xuewh22\}@mails.tsinghua.edu.cn, yuanly@ust.hk, wenping@tamu.edu,
    
    \{haopan, wangbins\}@tsinghua.edu.cn
}

\usepackage{bibentry}
\begin{document}

\maketitle

\begin{abstract}
Multi-view video reconstruction plays a vital role in computer vision, enabling applications in film production, virtual reality, and motion analysis. While recent advances such as 4D Gaussian Splatting (4DGS) have demonstrated impressive capabilities in dynamic scene reconstruction, they typically rely on the assumption that input video streams are temporally synchronized. However, in real-world scenarios, this assumption often fails due to factors like camera trigger delays or independent recording setups—leading to temporal misalignment across views and reduced reconstruction quality. To address this challenge, a novel temporal alignment strategy is proposed for high-quality 4DGS reconstruction from unsynchronized multi-view videos. Our method features a coarse-to-fine alignment module that estimates and compensates for each camera's time shift. The method first determines a coarse, frame-level offset and then refines it to achieve sub-frame accuracy. This strategy can be integrated as a readily integrable module into existing 4DGS frameworks, enhancing their robustness when handling asynchronous data. Experiments show that our approach effectively processes temporally misaligned videos and significantly enhances baseline methods.
\end{abstract}


\section{Introduction}

The reconstruction of dynamic scenes in four dimensions (4D) is a frontier topic in computer graphics and computer vision, aiming to capture and reconstruct time-evolving 3D scenes from multi-view videos. This technology is fundamental to enabling applications such as free-viewpoint video, immersive virtual and augmented reality experiences, digital twins, and visual effects production. In recent years, 4D Gaussian Splatting (4DGS) has emerged as a leading paradigm in dynamic scene reconstruction, distinguished by its exceptional rendering quality and unprecedented real-time rendering speeds. By representing the scene with explicit Gaussian primitives, it achieves highly efficient and high-fidelity modeling of complex dynamic details.

However, a critical yet demanding assumption underpinning most state-of-the-art 4DGS methods~\cite{scgs,yang2023gs4d,wu20244d} is the requirement for strict temporal synchronization across all capture cameras. This implies that for any given timestamp $t$, the camera shutters are triggered simultaneously. While this condition can be met in professional studios or laboratory settings using expensive hardware like Genlock signal generators, this assumption is often violated in more general, real-world capture scenarios. For instance, when using a set of independent consumer-grade cameras (e.g., smartphones, GoPros) or a distributed camera system controlled over a wireless network, the lack of a centralized clock signal, coupled with network latency and manual start-and-stop operations, almost inevitably introduces temporal misalignment ranging from milliseconds to even seconds between video streams.
\begin{figure}[]
    \centering
    \includegraphics[width=1.0\linewidth]{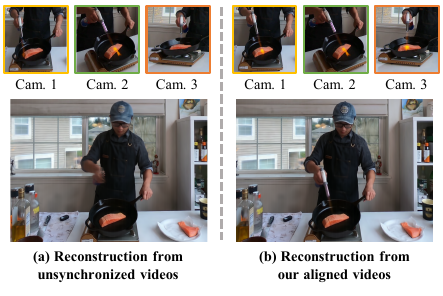}
    \caption{We introduce a novel readily integrable module that significantly enhances existing 4D dynamic scene reconstruction methods (e.g., 4DGaussians and SC-GS). Compared to the baseline result (left), our method (right) demonstrates superior quality in capturing intricate details and complex motion.}
    \label{fig:teaser}
\end{figure}
This temporal asynchronicity poses a critical challenge to 4D reconstruction. When a model attempts to fuse views captured at physically distinct moments to reconstruct the scene at a single logical timestamp, severe visual artifacts arise. For objects in fast motion, even minor time discrepancies can cause significant positional differences between views, leading to ghosting artifacts and motion blur in the final reconstruction. This inconsistent observational data misguides the 4DGS optimization process, causing it to incorrectly attribute temporal errors to flaws in spatial geometry or appearance. This ultimately leads to a sharp degradation in reconstruction quality, or even complete failure. Consequently, the problem of temporal misalignment has become a key bottleneck preventing the widespread adoption of high-fidelity 4D reconstruction technology beyond controlled environments.

To address this challenge, we propose a novel and general temporal alignment strategy for high-quality 4DGS reconstruction from unsynchronized multi-view videos. Instead of passively accepting data with temporal errors, we explicitly incorporate the unknown temporal offset of each camera into the optimization objective. Specifically, we design a coarse-to-fine optimization framework to estimate and compensate for each camera's time shift precisely. The framework first operates on a coarse-grained temporal scale to rapidly identify an approximate offset range, effectively preventing the optimization from converging to a poor local minimum. Subsequently, it performs fine-tuning at a more granular scale to achieve sub-frame alignment accuracy. Our strategy can be seamlessly integrated as a readily integrable module into various existing 4DGS frameworks, significantly enhancing their robustness and reconstruction quality when handling asynchronous data.

The main contributions of this paper are summarized as follows:
\begin{itemize}
    \item We systematically identify and analyze the problem of temporal misalignment, which is prevalent in real-world capture, and reveal its severely detrimental effects on existing 4DGS methods.
    \item We propose, for the first time, a coarse-to-fine optimization framework that jointly solves for the unknown temporal offsets of each camera stream concurrently with the 4DGS reconstruction process.
    \item Extensive experiments demonstrate that our method effectively handles temporally asynchronous videos, significantly outperforms baseline approaches on multi-view datasets, and successfully reconstructs high-quality, artifact-free dynamic scenes. This work substantially expands the application boundaries of dynamic Gaussian reconstruction, enabling its use with lower-cost and more flexible capture setups.
\end{itemize}

\section{Related Work}
\subsection{Novel View Synthesis}

Novel view synthesis (NVS) for dynamic scenes has advanced considerably in recent years, fueled by breakthroughs in Neural Radiance Fields (NeRF) and 3D Gaussian Splatting (3DGS).
The foundational work, NeRF~\cite{mildenhall2021nerf}, introduced a method to represent 3D scenes as continuous volumetric functions parameterized by neural networks. Various strategies have been proposed to reduce memory consumption and speed up the reconstruction process in neural rendering. These include techniques such as hash-based encoding~\cite{muller2022instantNGP}, mobile-optimized architectures~\cite{chen2022mobilenerf}, scalable training frameworks~\cite{wu2022scalable}, highly parallelized networks~\cite{reiser2021kilonerf}, and efficient factorized representations~\cite{garbin2021fastnerf}, all of which contribute to making real-time or large-scale neural rendering more feasible.

The advent of 3DGS marked a shift towards rasterization-based explicit methods. 3DGS~\cite{kerbl20233d} modeled scenes as collections of 3D Gaussians, enabling real-time rendering. Following the introduction of 3DGS, extensive research has focused on improving its storage efficiency, rendering speed, and visual quality.
Scaffold-GS~\cite{scaffoldgs} improves 3DGS by replacing unstructured Gaussians with a structured scaffold of anchors that generate view-dependent Gaussians, effectively reducing redundancy and enhancing view consistency.
The COLMAP-free approach~\cite{colamp-free-3dgs} enables the progressive reconstruction of a 3D Gaussian scene by simultaneously estimating camera poses from video input.
GaussianPro~\cite{cheng2024gaussianpro} introduces a progressive propagation strategy that efficiently densifies the Gaussian cloud from a sparse initialization to capture finer details.
To optimize rendering performance and consistency, Octree-GS~\cite{ren2024octree} organizes Gaussians into a level-of-detail (LOD) hierarchy using an octree structure, ensuring real-time rates at different viewing distances.
Departing from the standard primitive, GES~\cite{hamdi2024ges} proposes a generalized exponential function as a more flexible alternative to the 3D Gaussian, improving rendering efficiency and quality.
In parallel, extensions of the original method have also been explored, including approaches~\cite{huang20242d} that leverage 2D Gaussian primitives to achieve smoother surface representations.

\begin{figure*}[t]
    \centering
    \includegraphics[width=\linewidth]{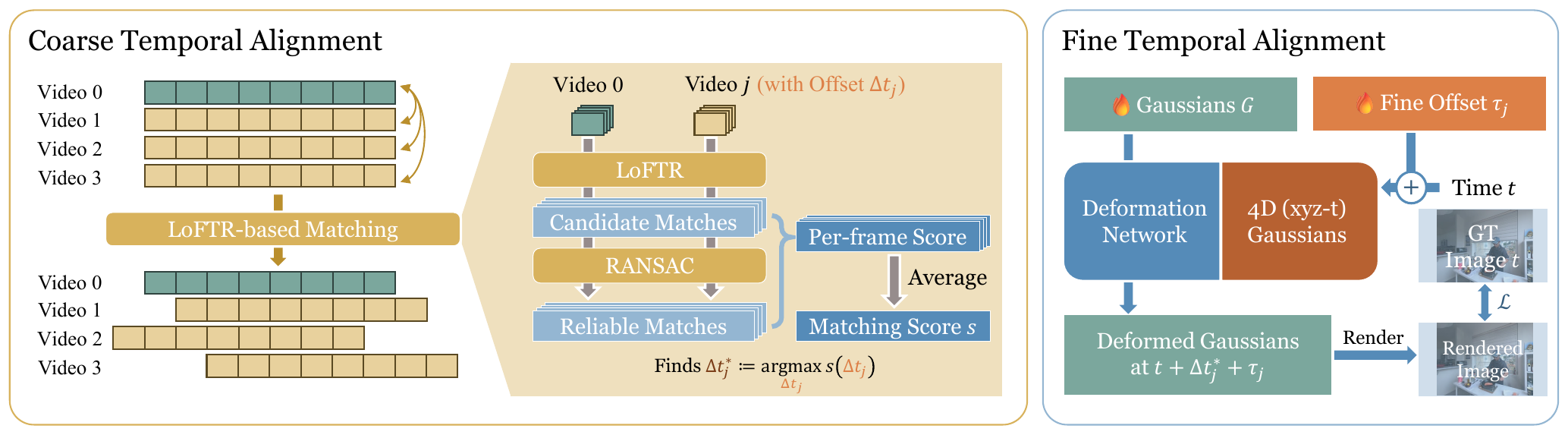}
    \caption{Overview of our two-stage temporal alignment pipeline. \textbf{Left}: Coarse Temporal Alignment estimates integer offsets \(\Delta t_j^*\) by matching frames across videos using LoFTR and RANSAC. \textbf{Right}: Fine Temporal Alignment refines offsets with a learnable \(\tau_j\). The result is supervised by a photometric reconstruction loss.}
    \label{fig:pipeline}
\end{figure*}

\subsection{Dynamic Gaussian Reconstruction}

Recent advancements have extended 3D Gaussian Splatting to dynamic scenes.
Deformable 3DGS~\cite{yang2024deformable} introduces a deformable formulation that learns scene representations in a canonical space, using a deformation field to capture monocular dynamics.
4DGaussians~\cite{wu20244d} proposes a hybrid explicit representation that combines 3D Gaussians with 4D neural voxels, employing decomposed voxel encoding and a lightweight MLP to predict Gaussian deformations over time.
Building on these foundations, subsequent works~\cite{duan20244d,yan20244d,yang20244d} have improved dynamic modeling through enhancements in primitive representation, motion optimization, and sampling strategies.
SplineGS~\cite{park2024splinegs} further models smooth Gaussian trajectories using cubic Hermite splines with motion-adaptive control point pruning, enabling efficient and expressive deformation under varying motion patterns.

Recent methods for dynamic reconstruction from monocular real-world videos can be categorized into reconstruction-based approaches, which explicitly recover scene geometry and motion over time~\cite{liu2025modgs,wang2024shape,stearns2024dynamic,lei2024mosca}, and prediction-based approaches, which directly estimate motion or deformation using priors or flow-based representations~\cite{wang2024gflow,liang2024feed}.
These methods collectively advance the accuracy, coherence, and efficiency of dynamic scene understanding in real-world settings.
GaussianFlow~\cite{lin2024gaussian} and MotionGS~\cite{zhu2024motiongs} incorporate optical flow information to introduce additional motion constraints in dynamic motion modeling, significantly improving temporal consistency and geometric accuracy.

\section{Preliminary}
\paragraph{3D Gaussian Splatting} In the 3D Gaussian Splatting (3DGS) framework, a scene is modeled as a collection of anisotropic 3D Gaussians $\{G_i\}$. Each Gaussian \(G\) is parameterized by its center \(\mu\in\mathbb{R}^3\), covariance matrix \(\Sigma\in\mathbb{R}^{3\times3}\), opacity \(\alpha\in\mathbb{R}\), and color \(c\in\mathbb{R}^3\). After scaling by opacity, its density at any point \(\mathbf{x}\in\mathbb{R}^3\) is given by
\begin{equation}
        G(\mathbf{x})=\alpha\exp\left[-\frac12\,(\mathbf{x}-\mu)^\top\Sigma^{-1}(\mathbf{x}-\mu)\right].
\end{equation}
Its covariance matrix \(\Sigma\) is parameterized as
\begin{equation}
     \Sigma = R S S^\top R^\top
\end{equation}
where \(R\in\mathrm{SO}(3)\) is the rotation matrix and \(S=\mathrm{diag}(\mathbf{s})\) is a diagonal scale matrix.
Given a camera’s extrinsic matrix \(P\) and intrinsic matrix \(K\), each 3D Gaussian is projected onto the image plane using an affine approximation \(W\) of perspective projection \(P\). The projected 2D Gaussian has parameters
\begin{equation}
    \mu' = \Pi\left(KW\mu\right),
    \quad
    \Sigma' = JW\Sigma W^\top J^\top
\end{equation}
where \(\Pi\) denotes the standard perspective division and \(J\) is the Jacobian of the affine approximation to the full nonlinear projection. Finally, these 2D Gaussians are composited via alpha blending to produce both RGB images and depth maps in a single, efficient rasterization pass.

The input for 3DGS scene reconstruction generally involves a set of posed images, and the 3DGS model is optimized to minimize the difference between its rendered images and the given images at the given poses. 

\paragraph{4D Gaussian Splatting}
4DGS extends the static scene of 3DGS with an extra temporal dimension. This is typically achieved by associating each 3D Gaussian primitive with the time dimension, i.e. $G(\mathbf{x})$ becomes $G(\mathbf{x},t)$, where the parameters of a Gaussian primitive can vary with $t$~\cite{wu20244d, yang2023gs4d}.

The input for 4DGS scene reconstruction involves a set of videos rather than static images, where the videos do not have to be synchronized in the time dimension in practice, and the 4DGS model is reconstructed by minimizing the difference between its rendered images from the given videos.

The unsynchronized videos pose additional challenge for 4DGS reconstruction, because for the 4DGS model $\{G_i(\mathbf{x},t)\}$ at the moment $t$, it's not clear to which frame $I_{j,t'}$ of the $j$-th video the GS model should compare.
As we can see in Fig.~\ref{fig:teaser}, a naive treatment of $t'=t$ leads to severe reconstruction failures, especially for highly dynamic regions.
Our goal in this paper is to present an approach such that an optimal temporal alignment $t'=f_j(t)$ is found for each video, so that the total reconstruction error is minimized.
In this paper, we assume the temporal alignment function $f_j(t)=t+\delta t_j$ is an offset transformation, which complies with practical scenarios where the videos are unsynchronized mostly by a time shift.

\section{Method}

\begin{figure*}[htbp]
    \centering
    \includegraphics[width=0.95\textwidth]{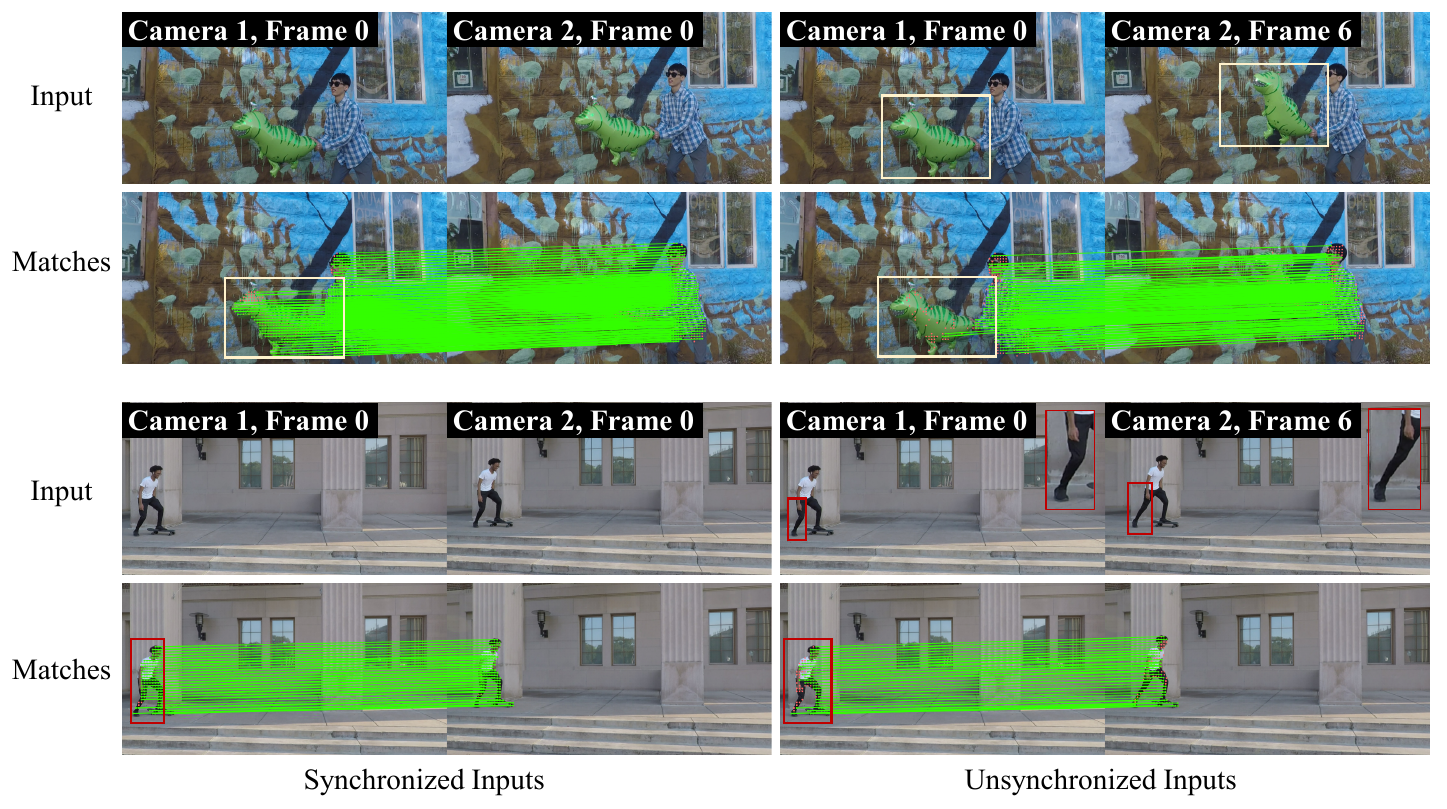}
    \caption{Illustration of our coarse temporal alignment. For each candidate temporal offset $\Delta t$ between two camera views, we evaluate the number of geometrically consistent feature matches, quantified by the number of RANSAC inliers. For a given view pair, we first extract putative matches using LoFTR (red dots), then apply RANSAC to robustly find inliners (green lines), which serve as our alignment score.}
    \label{fig:time-align}
\end{figure*}

We present a time alignment module in our method, designed to address the challenge of reconstructing dynamic scenes from asynchronous multi-view videos. Rather than proposing an entirely new reconstruction pipeline, our key contribution lies in a modular and pluggable component that can be seamlessly integrated into existing state-of-the-art dynamic Gaussian methods. The module is intended to automatically estimate and compensate for temporal misalignments caused by hardware limitations or capture inconsistencies in multi-camera systems.

In the following sections, we first introduce our coarse-to-fine time alignment module, where the time shift of each camera $\delta t = \Delta t + \tau$ is decomposed into a coarse level shift $\Delta t$ and a fine level shift $\tau$ to assist globally optimal alignment. Then we show how our module can be easily plugged into state-of-the-art dynamic representations implementing 4DGS.

\subsection{Coarse Temporal Alignment}

To address temporal asynchrony among multi-view video streams, we first perform a coarse, frame-level synchronization. Our approach is based on the insight that dense feature matchers like LoFTR~\cite{sun2021loftr}, while designed for static scenes, can be powerfully repurposed for temporal alignment. Specifically, when video frames from different views are captured at the exact same moment, a dynamic foreground effectively becomes static from multiple perspectives. For LoFTR, this "instantaneously static" object provides a rich source of stable features, leading to a surge in the number of high-confidence matches, as illustrated by Fig.~\ref{fig:time-align}. We leverage this principle by identifying the frame pairings that maximize these feature matches, thereby establishing a robust initial synchronization.

Building on this insight, we cast temporal alignment as a frame-pair selection problem: for each pair of camera views, we search for the time offset $\Delta t$ that matches the views best,
as measured by the number of reliable feature correspondences.

\begin{figure}[htbp]
    \centering
    \includegraphics[width=1.0\linewidth]{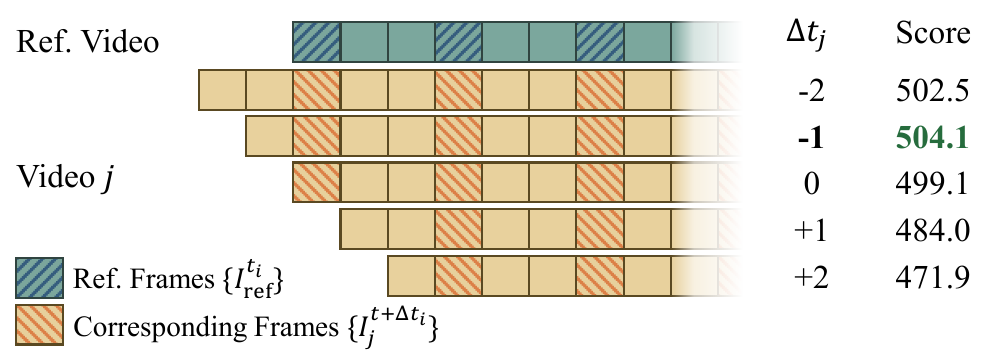}
    \caption{Corresponding frames for matching score calculation under different \(\Delta t_j\)s. We find the offset \(\Delta t_j\) for each video \(j\) relative to the reference video that maximizes the matching score.}
    \label{fig:coarse-matching}
\end{figure}

As a preliminary step, we process all video streams through a pre-trained video segmentation model~\cite{yang2023track} to obtain a binary foreground mask, $M$, for every frame. This mask, $M^t$, isolates the pixels corresponding to the dynamic subject in frame $I^t$, By focusing exclusively on the dynamic subject, this foreground-centric strategy provides a more direct and reliable signal for synchronization.

For a set of multi-view videos as the input, we select one as the reference, from which reference frames \(\{I_{\text{ref}}^{t_i}\}\) are sampled. For each other video \(j\), we search for the best \(\Delta t_j\) in the range \([-k, k]\), where \(k\) is a hyperparameter to account for the maximum potential misalignment.

For every candidate \(\Delta t_j\), we calculate the average matching score between each reference frame \(I_\text{ref}^{t_i}\) and its corresponding frame \(I_j^{t_i+\Delta t_j}\). The correspondence is illustrated in Fig.~\ref{fig:coarse-matching}. The score is computed in two stages:
generating candidate correspondences, and then estimating robust geometric matches.

In the first stage, we use LoFTR to search for candidate feature correspondences between each frame pair $(I_\text{ref}^t, I_{j}^{t+\Delta t_j})$. Critically, we filter those correspondences using the foreground masks, retaining only those matches where both points lie within their respective masks, $M_\text{ref}^t$ and $M_{j}^{t+\Delta t_j}$.

In the second stage, we apply the RANSAC~\cite{ransac} algorithm to this set of foreground-only correspondences. RANSAC robustly fits a geometric model (in our case, a fundamental matrix) to find the largest set of self-consistent inliers. This step effectively tests the hypothesis that the foreground object captured in the two views is a rigid projection from a single moment in time.

The number of foreground inliers, $N_{\text{inlier-fg}}(I_\text{ref}^{t_i},I_j^{t_i+\Delta t})$, serves as the alignment score. A high score indicates that the subject's pose is geometrically consistent between the two views, providing strong evidence of correct temporal alignment. The optimal offset $\Delta t^*_j$ for video \(j\) is the one that maximizes this score:
\begin{equation}
    \Delta t^*_j = \argmax_{\Delta t_j \in [-k, k]} \sum_{t_i}{N_{\text{inlier-fg}}(I_\text{ref}^{t_i},I_j^{t_i+\Delta t})}.
\end{equation}

\subsection{Fine Temporal Refinement}
The coarse alignment provides a robust, integer-frame synchronization for all video streams. These offsets are added to the frame indices of each source camera stream $j$.
To achieve sub-frame precision, we introduce a small, learnable temporal refinement parameter, $\tau_j$, for each camera $j$. This parameter represents a continuous, residual time shift that is learned during training. $\tau_j$, in addition to $\Delta t_j^*$, is added to the timestamp before calculating the parameters of Gaussians at time \(t\), resulting in the queried \(t'\) being 
\begin{equation}
    t' = t + \Delta t^*_j + \tau_j\,.
\end{equation}
These refinement parameters $\{\tau_j\}$ are learnable and are optimized jointly with the 4DGS model. The gradients are derived from the end-to-end photometric reconstruction loss. This allows the optimization process to dynamically discover and correct any residual, sub-frame temporal discrepancies, which is essential for minimizing motion artifacts and achieving high-fidelity dynamic scene reconstruction.

\subsection{Integration with 4D Representations}

Our coarse alignment module is an individual part that does not require modifying the baseline implementation. The fine temporal alignment needs to be implemented into existing methods. 

\subsubsection{Neural 4D Representations}
\label{sec:neural-based}

The proposed module can be integrated with 4D Gaussian Splatting systems that represent scene dynamics through neural networks---a paradigm widely adopted by recent high-performance dynamic 4DGS methods~\cite{wu20244d,scgs}. 
These systems model dynamics by using a neural network to deform a static set of canonical 3D Gaussians.

The neural deformation network $\mathcal{D}_\theta$, parameterized by $\theta$ and typically implemented as a multi-layer perceptron (MLP), models the temporal evolution of each Gaussian.
Specifically, it takes as input the positionally encoded canonical location $\gamma(\bm{\mu}_k)$ and time $\gamma(t)$, and outputs deformations, such as position offsets $\Delta\bm{\mu}_k$:
\begin{equation}
    \mathcal{D}_\theta(\gamma(\bm{\mu}_k), \gamma(t)) \rightarrow (\Delta \bm{\mu}_k, \dots)
\end{equation}
where $\gamma(\cdot)$ denotes a positional encoding function.

Our module interfaces with the deformation network via its temporal input. Instead of feeding the global time $t$, we add the coarse offset \(\Delta_j^*\) and a per-camera learnable offset \(\tau_j\) to each frame.
Since the output of $\mathcal{D}_\theta$ is differentiable with respect to its temporal input, gradients with respect to the per-camera offset $\tau_j$
can be naturally computed during backpropagation, thereby enabling end-to-end optimization.

\subsubsection{Direct 4D Representations}

An alternative approach to representing dynamic scenes is to extend 3D Gaussians directly into a 4D (xyz-t) space. The RT4DGS method~\cite{yang2023gs4d} adapts the 3DGS methodology by parameterizing 4D covariance matrices using 4D rotations and 4D scales.

Similar to 3DGS, rendered images from this representation are fully differentiable to their inputs. However, the original RT4DGS implementation does not include gradient computation concerning the input timestamp $t$. Therefore, we compute the derivative of the reconstruction loss \(\mathcal L\) with respect to $t$ using a finite difference approximation:
\begin{equation}
\frac{\partial \mathcal L}{\partial t} \approx \frac{\mathcal L(t+h) - \mathcal L(t)}{h}.
\label{eq:finite_difference}
\end{equation}

The selection of an appropriate step size, \(h\), is crucial for numerical stability. In our setup, we find that values for \(h\) ranging from one-hundredth to one-tenth of the frame interval provide stable results.

\section{Experiments}
\begin{table*}[htbp]
    \centering
    \begin{tabular*}{\textwidth}{l @{\extracolsep{\fill}} ccc ccc ccc}
        \toprule
        \multirow{2}{*}{\textbf{Method}} & \multicolumn{3}{c}{\textbf{Coffee Martini}} & \multicolumn{3}{c}{\textbf{Cook Spinach}} & \multicolumn{3}{c}{\textbf{Cut Roasted Beef}} \\
        \cmidrule(r){2-4} \cmidrule(lr){5-7} \cmidrule(l){8-10}
        & PSNR$\uparrow$ & SSIM$\uparrow$ & LPIPS$\downarrow$ & PSNR$\uparrow$ & SSIM$\uparrow$ & LPIPS$\downarrow$ & PSNR$\uparrow$ & SSIM$\uparrow$ & LPIPS$\downarrow$ \\
        \midrule
        Sync-Nerf & 27.64 & 0.893 & 0.147 & 28.90 & 0.918 & 0.132 & 29.96 & 0.925 & 0.125 \\
       \midrule
        SC-GS~\cite{scgs}     &  \textbf{26.81}     &  \textbf{0.906}     &  \textbf{0.113}     &  30.53     &   0.939    & 0.104      & 31.13      &  0.943     &   0.095    \\
        SC-GS+Ours& 25.68      &  0.898   & 0.125     & \textbf{31.54}     &   \textbf{0.947 }   &   \textbf{0.089}     &  \textbf{31.37}     &  \textbf{0.949}     & \textbf{0.084}        \\
       
        \midrule
        4DGaussians~\cite{wu20244d}      & 26.44 & 0.905 &  0.120     &  31.44     &  0.946      & 0.098      & 31.37    & 0.941      & 0.096      \\
        4DGaussians+Ours &  \textbf{28.01}     &  \textbf{0.918}   &  \textbf{0.108}     &   \textbf{32.57}    &   \textbf{0.951}    &   \textbf{0.089}    & \textbf{31.71}    &   \textbf{0.948}    &  \textbf{0.093}     \\
        \midrule
        RT4DGS{*}~\cite{yang2023gs4d}     &  27.92      & 0.919      & 0.085      &    31.15   & 0.948      & 0.077      & 31.43      & 0.953      & 0.072      \\
        RT4DGS{*}+Ours      & \textbf{28.35}       &  \textbf{0.924}     &  \textbf{0.079}    &  \textbf{33.15}     & \textbf{0.962}      & \textbf{0.059}      & \textbf{32.94}     & \textbf{0.963}      & \textbf{0.056}      \\
        \midrule[\heavyrulewidth]
        \multirow{2}{*}{\textbf{Method}} & \multicolumn{3}{c}{\textbf{Flame Salmon}} & \multicolumn{3}{c}{\textbf{Flame Steak}} & \multicolumn{3}{c}{\textbf{Sear Steak}} \\
        \cmidrule(r){2-4} \cmidrule(lr){5-7} \cmidrule(l){8-10}
        & PSNR$\uparrow$ & SSIM$\uparrow$ & LPIPS$\downarrow$ & PSNR$\uparrow$ & SSIM$\uparrow$ & LPIPS$\downarrow$ & PSNR$\uparrow$ & SSIM$\uparrow$ & LPIPS$\downarrow$ \\
        \midrule
        Sync-Nerf & 27.00 & 0.890 & 0.150 & 30.66 & 0.942 & 0.093 & 30.50 & 0.938 & 0.110 \\
        \midrule
        SC-GS~\cite{scgs}     &  26.83     &   0.912    &  0.105     &   30.34    &   0.947    & 0.096       &  30.92     &  \textbf{0.950}     &   0.085    \\
        SC-GS+Ours&   \textbf{27.10}   & \textbf{0.915}      & \textbf{0.102}      &    \textbf{31.46}   &  \textbf{0.953}     & \textbf{0.081}      &    \textbf{31.20}   &   0.949    &  \textbf{0.092}     \\
       
        \midrule
        4DGaussians~\cite{wu20244d}      &  28.01      & 0.917      & 0.109      & 30.68       &  0.952      &  0.087     &    29.67    &  0.947      &   0.082    \\
        4DGaussians+Ours &  \textbf{29.53}     &  \textbf{0.923}    &  \textbf{0.103}     &  \textbf{32.63}     &  \textbf{0.955}     &   \textbf{0.077}   &    \textbf{32.51}   &   \textbf{0.959}    &    \textbf{0.080}    \\
        \midrule
         RT4DGS{*}~\cite{yang2023gs4d}     &  27.78     & 0.923      & 0.080      & 31.13      & 0.956       & 0.068       & 32.94      & 0.965      & 0.058      \\
        RT4DGS{*}+Ours      &   \textbf{28.79}    & \textbf{0.929}      & \textbf{0.070}      &  \textbf{33.34}     & \textbf{0.968}     & \textbf{0.050}      &    \textbf{33.51}   & \textbf{0.968}      & \textbf{0.055}      \\
        
        \bottomrule
    \end{tabular*}
    \caption{Quantitative comparison of our method against baselines on the Dynerf dataset. Our method consistently enhances RT4DGS, 4DGaussians, and SC-GS frameworks. RT4DGS*: We follow RT4DGS using half the resolution for evaluation.}
    \label{tab:quantitative_results}
\end{table*}

\begin{figure*}[htbp]
    \centering
    \includegraphics[width=1.0\textwidth]{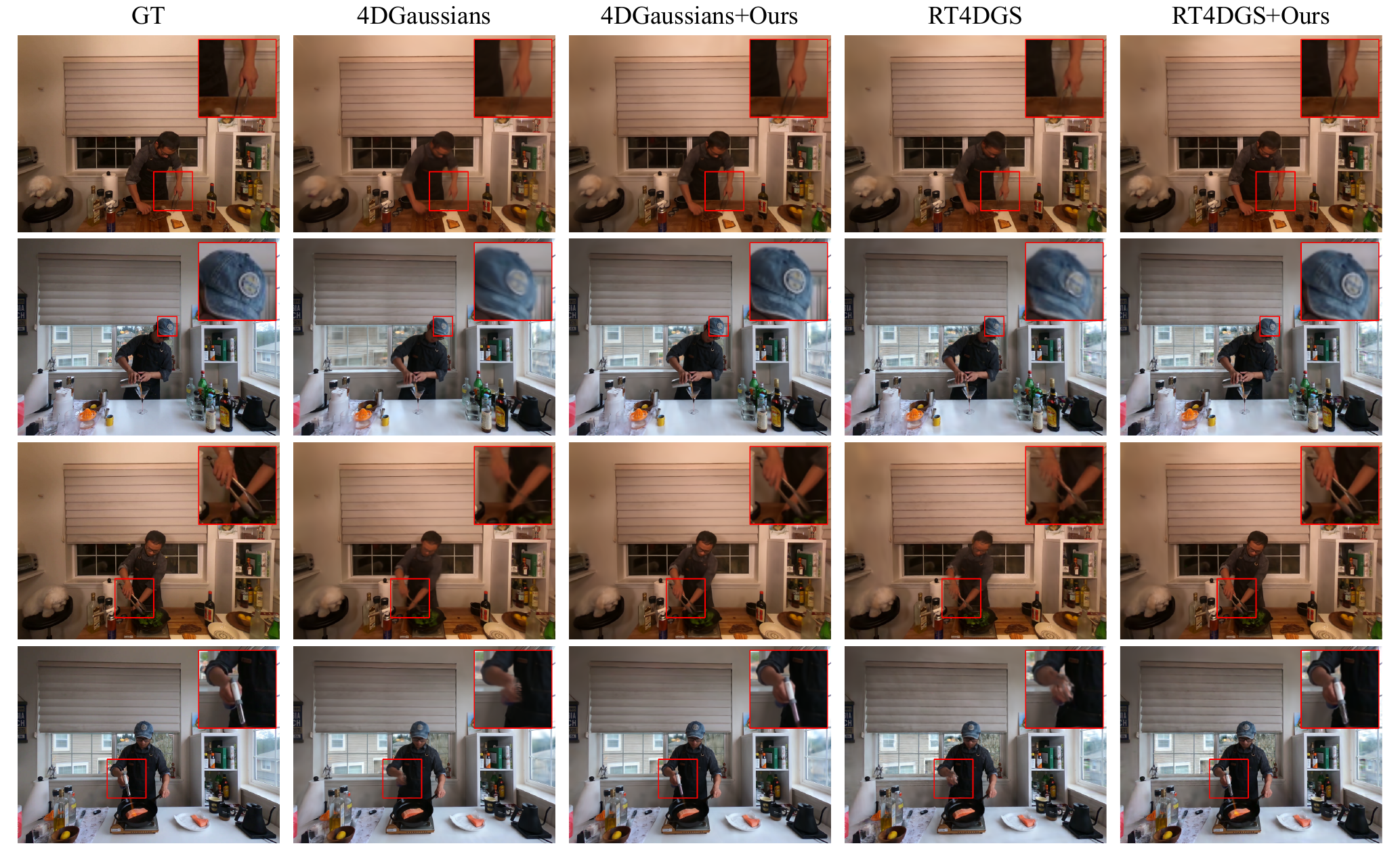}
    \caption{\textbf{Visual comparison of reconstruction results from unsynchronized inputs.} We compare novel view synthesis results from the original 4DGaussians and RT4DGS methods against versions enhanced by our approach (+Ours). Images are from four scenes of the DyNeRF dataset: coffee\_martini, cut\_roasted\_beef, cook\_spinach, and flame\_salmon. }
    \label{fig:result_compare}
\end{figure*}

\subsection{Datasets and Metrics}
To evaluate our method, we utilize the Dynerf dataset~\cite{neu3d}, which contains six challenging dynamic scenes, each captured from approximately 20 viewpoints in 9-second video clips. To heighten the challenge and better simulate real-world conditions, we create a more demanding benchmark from this data. The videos are subsampled to 15 FPS, which introduces more significant motion between consecutive frames. We apply a random temporal offset of up to 10 frames to each video. This process yields a multi-view, temporally misaligned dataset that mimics asynchronous capture scenarios.
On this benchmark, we assess our method's performance on the task of novel view synthesis. We report three standard quantitative metrics: PSNR, SSIM, and LPIPS~\cite{zhang2018unreasonable}.

\subsection{Implementation Details}
Experiments based on 4DGaussians and SC‑GS were conducted on an RTX 3090, while those using the RT4DGS method were run on an RTX A6000. Initial point clouds were generated using the COLMAP SfM pipeline.
For fair comparison, the hyperparameters from baseline methods are untouched. We choose \(h\) to be one-thirtieth of the frame interval for derivative calculation.

\subsection{Comparisons}

We integrate our module into three state-of-the-art 4D scene reconstruction pipelines: 4DGaussians~\cite{wu20244d}, SC-GS~\cite{scgs}, and RT4DGS~\cite{yang2023gs4d}. Direct comparisons with the original baselines demonstrate the effectiveness of our approach. As shown in Tab.~\ref{tab:quantitative_results}, our method achieves significant quantitative improvements. Qualitative results in Fig.~\ref{fig:result_compare} further illustrate reduced visual artifacts, particularly those caused by unsynchronized reconstruction inputs.

\subsection{Ablation Study}

To analyze the contribution of each component within our proposed temporal refinement module, we conduct a series of ablation studies. These experiments are designed to validate two key aspects of our method: (1) the necessity of our coarse-to-fine strategy for predicting time offsets, and (2) the robustness of our model against varying degrees of random temporal perturbations.

\paragraph{Ablation on Coarse and Fine Temporal Alignment}
We investigate the effectiveness of our proposed two-stage (Coarse + Fine) strategy for time offset prediction. To this end, we compare our \textbf{Full} method against three variants: a baseline without any temporal correction, a \textbf{Coarse}-only version, and a \textbf{Fine}-only version. Quantitative results are provided in Tab.~\ref{tab:ablation_method}.

\begin{figure}[!htbp]
    \centering
    \includegraphics[width=1.0\linewidth]{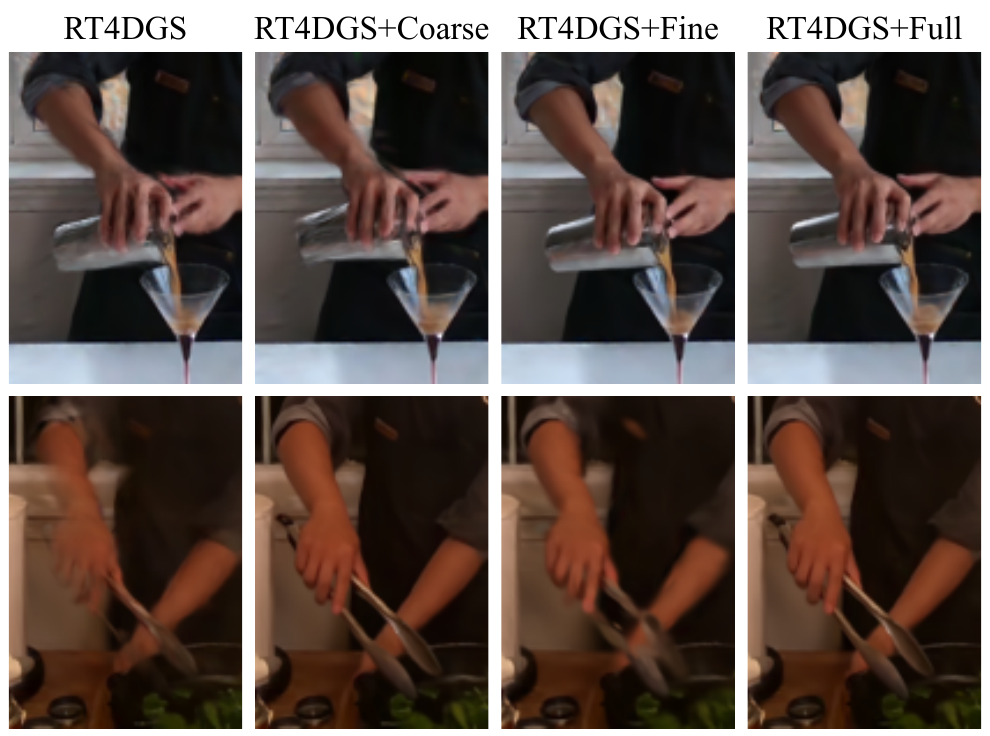}
    \caption{Ablation study results. The full method yields the best reconstruction quality, while removing components leads to visible performance drops.}
    \label{fig:ablation_method}
\end{figure}

As shown in Fig.~\ref{fig:ablation_method}, both the Coarse-only and Fine-only models significantly reduce visual artifacts such as blurring and ghosting compared to the baseline. This demonstrates that each stage independently contributes to mitigating temporal misalignment.
However, the Coarse-only or the Fine-only results still suffer from residual misalignment in dynamic scenes.
In contrast, the full two-stage approach achieves the best visual and quantitative results.
Both the quantitative metrics in Tab.~\ref{tab:ablation_method} and the qualitative comparisons in Fig.~\ref{fig:ablation_method} confirm that the combined Coarse + Fine strategy is essential for achieving optimal reconstruction quality.

\begin{table}[h]
\centering
\begin{tabular}{l|ccc}
\hline
\textbf{Method} & \textbf{PSNR}$\uparrow$ & \textbf{SSIM}$\uparrow$ & \textbf{LPIPS}$\downarrow$ \\
\hline
4DGaussians  & 29.56 & 0.935 & 0.099 \\
4DGaussians+Coarse     & 30.92 & 0.943          & 0.092 \\
4DGaussians+Fine       & 30.87 & 0.941          & \textbf{0.091} \\
4DGaussians+Full       & \textbf{31.16} &  \textbf{0.942} & \textbf{0.091} \\
\hline
\end{tabular}
\caption{Our full two-stage method (4DGaussians+Full) achieves the best performance, highlighting the complementary roles of coarse and fine temporal refinement.}
\label{tab:ablation_method}
\end{table}

\paragraph{Ablation on Random Time Offsets}
To assess the robustness and performance of our method under varying levels of input asynchronicity, we conduct an ablation study by introducing different magnitudes of random time offsets. During training, we add a temporal jitter $\Delta t$ to the timestamp of each input view, where $\Delta t$ is sampled uniformly from the range $[0, \tau_{\text{max}}]$. We evaluate our model's performance by progressively increasing the upper bound of this jitter, setting the maximum offset $\tau_{\text{max}}$ to 3, 5, and 10 frames, respectively.

The experimental results in Tab.~\ref{tab:ablation_random} clearly show that the performance of the Baseline model degrades dramatically as the maximum offset $\tau_{\text{max}}$ increases.
In stark contrast, the method augmented with our temporal refinement module maintains high reconstruction fidelity even under the most severe perturbation.

\begin{table}[h]
\centering
\setlength{\tabcolsep}{4pt}
\begin{tabular}{c|ccc|ccc}
\hline
\multicolumn{1}{c|}{\multirow{2}{*}{\(\tau_{\text{max}}\)}} & \multicolumn{3}{c|}{4DGaussians} & \multicolumn{3}{c}{+Ours} \\
\multicolumn{1}{l|}{}                  & PSNR      & SSIM     & LPIPS     & PSNR   & SSIM   & LPIPS   \\ \hline
3                                      &       30.69    &  0.938        & 0.097          &    31.25     &   0.943      &   0.092       \\
5                                      &    30.31       &     0.938      &  0.097          & 31.29    &   0.943    &  0.091      \\
10                                     &   29.60        &    0.935      &    0.099       &   31.16     &    0.942    &     0.091    \\ \hline
\end{tabular}
\caption{Ablation study on random time offsets.}
\label{tab:ablation_random}
\end{table}

\section{Conclusion}
This paper addresses a prevalent yet challenging problem in computer vision and graphics: high-quality dynamic scene reconstruction from temporally unsynchronized multi-view videos. We introduce a novel temporal alignment module that, when integrated into existing 4D reconstruction frameworks, yields significant improvements in reconstruction quality. This was demonstrated on a custom benchmark created from the DyNeRF dataset, which was modified to include random temporal offsets to simulate real-world conditions. Ultimately, our work resolves the critical issue of temporal asynchrony, greatly expanding the application boundaries of high-quality 4D dynamic scene reconstruction. It enables high-fidelity dynamic capture using lower-cost, more flexible camera systems, thereby promoting the technology's widespread adoption for real-world applications beyond controlled environments.

\bibliography{aaai2026}

\end{document}